\documentclass[letterpaper]{article} 
\usepackage[submission]{aaai2026}  
\usepackage{times}  
\usepackage{helvet}  
\usepackage{courier}  
\usepackage[hyphens]{url}  
\usepackage{graphicx} 
\usepackage{natbib}  
\usepackage{caption} 
\usepackage{algorithm}
\usepackage{algorithmic}

\usepackage{booktabs} 
\usepackage{amsmath}
\usepackage{amssymb}
\usepackage{colortbl}
\usepackage{multirow}
\usepackage{newfloat}
\usepackage{listings}
\DeclareCaptionStyle{ruled}{labelfont=normalfont,labelsep=colon,strut=off} 
\floatstyle{ruled}
\newfloat{listing}{tb}{lst}{}
\floatname{listing}{Listing}
\makeatletter
\def\showauthors@on{T}
\def\copyright@on{T}
\makeatother
\title{LFTR: Learning-Free Token Reduction for Multimodal Large Language Models}
\author {
    Zihui Zhao\textsuperscript{\rm 1},
    Yingxin Li\textsuperscript{\rm 1},
    Yang Li\textsuperscript{\rm 1}
}
\affiliations{
  \textsuperscript{\rm 1}Tsinghua Shenzhen International Graduate School, Tsinghua University\\
  Shenzhen, China\\
  zzh23@mails.tsinghua.edu.cn, liyx23@mails.tsinghua.edu.cn, yangli@sz.tsinghua.edu.cn
}

\begin{document}

\maketitle

\begin{abstract}
Multimodal Large Language Models (MLLMs) have demonstrated exceptional success in various multimodal tasks, yet their deployment is frequently limited by substantial computational demands and prolonged inference times. Given that the vision modality typically contains more comprehensive information than the text modality, resulting in encoded representations comprising an extensive number of tokens, leading to significant computational overhead due to the quadratic complexity of the attention mechanism. Current token reduction methods are typically restricted to specific model architectures and often necessitate extensive retraining or fine-tuning, restricting their applicability to many state-of-the-art models. In this paper, we introduce a learning-free token reduction (LFTR) method designed for MLLMs. LFTR can be seamlessly integrated into most open-source MLLM architectures without requiring additional fine-tuning. By capitalizing on the redundancy in visual representations, our approach effectively reduces tokens while preserving the general inference performance of MLLMs. We conduct experiments on multiple MLLM architectures (LLaVA, MiniGPT, QwenVL), and our results show that LFTR achieves up to a $16\times$ reduction of visual tokens while maintaining or even enhancing performance on mainstream vision question-answering benchmarks, all in a learning-free setting. Additionally, LFTR is complementary to other acceleration techniques, such as vision encoder compression and post-training quantization, further promoting the efficient deployment of MLLMs. Our project is available at https://anonymous.4open.science/r/LFTR-AAAI-0528.
\end{abstract}


\section{Introduction}
\begin{figure}[htbp]
\centering
\includegraphics[scale=0.7]{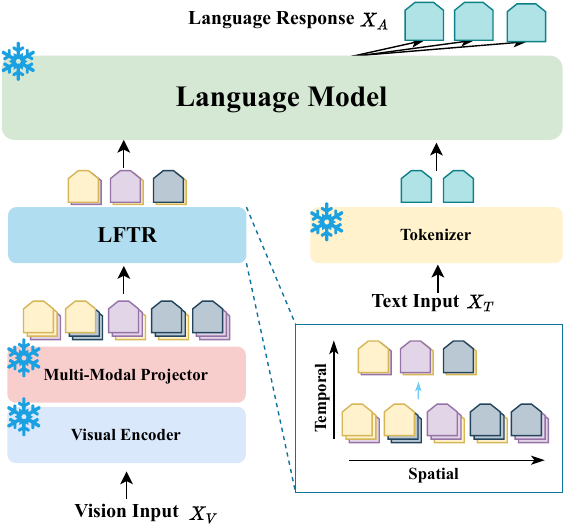}
\caption{Visual-LLM inference and our learning-free token reduction method. Our proposed method enables plug-and-play token reduction by utilizing only pretrained large language models. It achieves inference acceleration while preserving the inference capability of the model.}
\label{LLM}
\end{figure}

\label{intro}

Large Language Models (LLMs)\cite{achiam2023gpt,bai2023qwen,liu2024deepseek,jiang2024identifying,team2023gemini} have demonstrated remarkable performance in tasks such as text generation, translation, and comprehension. To extend these reasoning capabilities beyond text, Multimodal Large Language Models (MLLMs)\cite{liu2023visual,liu2024improved,lin2023video,ataallah2024minigpt4,chen2024far,zhang2024mm} incorporate multimodal encoders and projectors to align non-text modalities (e.g., visual) with language instructions. This design allows language backbone~\cite{touvron2023llama,touvron2023llama2,chiang2023vicuna} to leverage its pretrained knowledge for multimodal understanding without requiring extensive retraining. Among various modalities, vision stands out as a particularly promising direction due to its dominance in real-world information and its wide range of potential applications. Representative open source models such as Video-LLaVA~\cite{lin2023video} and MiniGPT4-Video~\cite{ataallah2024minigpt4} have demonstrated strong performance on a variety of downstream vision tasks.

Despite their promising capabilities, the widespread deployment of MLLMs remains hindered by substantial computational demands, particularly the high latency during inference. To address these challenges, recent efforts have focused on improving the efficiency of LLMs~\cite{jin2024efficient}. A variety of model compression techniques~\cite{chu2024mobilevlm,lin2024awq,zhang2023h2o,liu2024kivi} have been proposed to accelerate the inference process by optimizing the architecture of the LLM backbone. Beyond architectural compression, another major bottleneck arises from the large number of prefix tokens, especially visual tokens. This issue is particularly pronounced in dense visual modalities such as video, where the length of visual token sequences far exceeds that of textual inputs. The resulting increase in computational burden not only result in high time computational demands of model inference process but may also distract model from critical visual cues, thereby hindering its ability to perform accurate visual reasoning.

Observing the substantial redundancy among visual tokens, which often used as prefix inputs, several token reduction approaches have been introduced~\cite{bolya2023token,li2024tokenpacker,liang2022not,zhang2024token,shang2024llava,shi2024crossget}. While effective, most existing methods rely on fine-tuning or retraining of the MLLMs, which imposes practical constraints. For instance, fine-tuning large models such as LLaVA-1.5-13B~\cite{liu2024llavanext} demands extensive computational resources, requiring hundreds of GPU hours on A100 hardware. Moreover, many state-of-the-art (SOTA) MLLMs only release model architectures and parameters without disclosing the corresponding training data~\cite{beyer2024paligemma,wang2024qwen2}, rendering fine-tuning infeasible for the open source community. Previous works have explored learning-free approaches to reduce visual token counts during the inference stage of MLLMs (e.g., LLaVA-PruMerge~\cite{shang2024llava}, FreePruner~\cite{xu2024freepruner}). However, these methods are often tightly coupled with the underlying model architecture—particularly the design of the multimodal encoder—which can vary significantly across different MLLM frameworks. This architectural dependency limits the generalizability and practical applicability of such methods.

In this work, we present Learning-Free Token Reduction (LFTR) method, a universal and training-free framework for token reduction in vision-based multimodal language models. LFTR systematically identifies and removes redundant tokens, harnessing general importance estimation principles that are agnostic to model architecture. Our approach employs token significance metrics to preserve temporally important tokens and introduces a mathematically grounded, text-guided strategy for spatial token selection, ensuring that the retained tokens are both information-dense and contextually relevant to the task. By removing irrelevant or redundant visual content, our strategy can enhance the model inference ability for understanding and reasoning over visual representation. Our method can be easily integrated into popular MLLM structures with minimal engineering efforts and significantly improve the MLLM inference efficiency and reduce the computational cost.

We demonstrate that LFTR achieves up to $16\times$ compression of visual tokens, with either negligible or positive impact on downstream task performance, across a variety of MLLM architectures (LLaVA, MiniGPT, Qwen) and visual reasoning benchmarks. Our plug-and-play solution is fully compatible with existing acceleration techniques (such as quantization and model pruning), and does not require any model retraining or architectural modification.

Extensive experiments are conducted to validate the effectiveness of the proposed method in terms of both token reduction and inference performance. The main contributions are summarized as follows:

\begin{itemize}
\item We introduce a learning-free token reduction method for the MLLM inference process in a plug-and-play manner, which is adaptable to a wide range of MLLM architectures.

\item We propose four different token reduction strategies across temporal and spatial dimensions for video inputs, leveraging the intrinsic characteristics of visual modality to preserve significant information for language model inference.

\item Extensive experiments demonstrate that the proposed method significantly reduces token counts up to $16\times$ token reduction, accelerating MLLM inference while maintaining or even improving inference performance.
\end{itemize}

\section{Related Work}

\subsection{Multi-Modal Large Language Models}

Large Language Models, such as GPT-4~\cite{achiam2023gpt}, LLaMA~\cite{touvron2023llama,touvron2023llama2}, and Gemini~\cite{team2023gemini}, have demonstrated strong performance across a broad range of natural language understanding and generation tasks. Building on this success, recent advances in MLLMs have integrated vision encoders with language backbones to enable joint reasoning across text, images, audio, and video. Representative works such as CLIP~\cite{radford2021learning}, Flamingo~\cite{alayrac2022flamingo}, and BLIP-2~\cite{li2023blip} have significantly improved vision-language alignment. Open-source MLLMs, including LLaVA~\cite{liu2023visual}, MiniGPT-4~\cite{zhu2024minigpt}, and InternLM-XComposer~\cite{zhang2023internlm}, demonstrate strong instruction-following capabilities, while commercial systems such as GPT-4V~\cite{achiam2023gpt} and Gemini~\cite{team2023gemini} continue to advance the frontier of multimodal understanding.

However, these models often rely on a large number of visual tokens particularly when processing high resolution inputs which may result in substantial computational overhead during both training and inference. To alleviate this, several strategies have been proposed, including low-bit quantization~\cite{dettmers2022gpt3} and architectural compression~\cite{chu2023mobilevlm}. For example, TinyLLaVA~\cite{zhou2024tinyllava} investigates the impact of lightweight architectures and optimized training paradigms, achieving performance comparable to larger counterparts. Despite these efforts, the challenge of managing visual token redundancy remains a critical bottleneck in scaling efficient and deployable MLLMs.

\subsection{Token Reduction for LLMs}

Token reduction has become an essential technique for enhancing the computational efficiency of Transformer-based architectures, particularly in visual and multimodal tasks. Early approaches primarily focused on token pruning~\cite{rao2021dynamicvit,xu2022evo,liang2022not} and token merging~\cite{bolya2023token,marin2021token}, which aim to eliminate or consolidate redundant tokens to reduce the computational burden. More recent methods, such as DiffRate~\cite{chen2023diffrate} and PPT~\cite{wu2023ppt}, combine both pruning and merging within unified frameworks to adaptively reduce token redundancy during inference. 

Within the domain of MLLMs, visual token compression has drawn growing attention to address the prohibitive costs associated with high-resolution image and video inputs. Several recent methods have introduced modality-specific mechanisms to guide token selection. For instance, CrossGET~\cite{shi2024crossget} and MADTP~\cite{cao2024madtp} leverage alignment tokens to selectively retain informative visual representations.  Qwen-VL~\cite{wang2024qwen2} employs a resampler to downsample the visual token sequence to a fixed length, while FastV~\cite{chen2024image} implements an early-dropping strategy within the decoder layers to discard uninformative tokens, providing a lightweight yet effective alternative. LLaVA-PruMerge~\cite{shang2024llava} exploits spatial redundancy by measuring similarity between class and patch tokens to identify compressible regions using clustering. FreePruner~\cite{xu2024freepruner} proposes a strategy based on token contribution scores to retain tokens essential for high-level visual information.

While effective, these methods often rely on architectural modifications, customized fusion mechanisms, or fine-tuning, which can hinder their applicability in general-purpose or training-free settings. Such requirements pose significant barriers to deploying these techniques in real-world multimodal systems, where access to model internals or training data may be limited or entirely unavailable.

In contrast, we introduce a training-free token reduction method that requires neither model modifications nor supervision. Our approach presents a general, architecture-independent framework capable of achieving substantial compression while maintaining inference quality. By decoupling token reduction from model training, our method enables seamless integration into existing MLLM pipelines, offering a scalable and deployment-friendly solution to visual redundancy in multimodal processing.

\section{Preliminary}

Multimodal large language models(MLLMs) typically couple a vision encoder with a powerful language model backbone, enabling joint reasoning over visual and textual modalities. In this section, we formalize the multimodal tokenization and fusion process, focusing on image and video input scenarios which are most relevant to our work.

\subsection{Visual and Textual Encoding}

Let the visual input be an image or a video clip represented as a tensor $\mathbf{I} \in \mathbb{R}^{C \times H \times W}$ or $\mathbf{I}_{\text{vid}} \in \mathbb{R}^{N_t \times C \times H \times W}$, where $C$ is the number of channels, $H$ and $W$ are height and width, and $N_t$ is the number of video frames.

A widely adopted vision encoder in multimodal architectures is the Vision Transformer (ViT)~\cite{dosovitskiy2020image}, which divides each frame or image into a sequence of $N_s = \frac{H}{p} \times \frac{W}{p}$ non-overlapping patches of size $p \times p$, flattens each patch, and linearly projects it to a $d$-dimensional embedding. For video input $I_{vid}$, the set of patch tokens for frame $i$ is:
\begin{equation}
    \mathbf{Z}^{(i)}_V = \left[\mathbf{z}_{i,1}, \mathbf{z}_{i,2}, \ldots, \mathbf{z}_{i,N_s}\right] \in \mathbb{R}^{N_s \times d}
\end{equation}
where $\mathbf{z}_{i,p} \in \mathbb{R}^d$ is the embedding of the $p$-th patch in frame $i$.

For video inputs, each of the $N_t$ frames is independently encoded, producing a 3D tensor:
\begin{equation}
\mathbf{Z}_V = \left[\mathbf{Z}_V^{(1)}, \ldots, \mathbf{Z}_V^{(N_t)}\right] \in \mathbb{R}^{N_t \times N_s \times d}
\end{equation}
Some encoders(CLIP-ViT~\cite{radford2021learning}) may append a global [CLS] token per frame, denoted $\mathbf{z}_{i,\text{CLS}}$.

Given an input text prompt $X_T = (w_1, w_2, \dots, w_{L_T})$, a tokenizer and language embedding layer map it to a sequence of $L_T$ token embeddings:
\begin{equation}
\mathbf{Z}_T = \left[\mathbf{z}^{(T)}_1, \dots, \mathbf{z}^{(T)}_{L_T}\right] \in \mathbb{R}^{L_T \times d}
\end{equation}

\subsection{Multimodal Alignment and Reasoning}

In MLLM architectures, visual tokens are usually projected into the same embedding space as text tokens via a learnable linear layer (or MLP) $f_P$:
\begin{equation}
\tilde{\mathbf{z}}_{i,p} = f_P(\mathbf{z}_{i,p}), \quad \forall i=1,\ldots,N_t,\, p=1,\ldots,N_s
\end{equation}
yielding projected visual tokens:
\begin{equation}
\tilde{\mathbf{Z}}_V \in \mathbb{R}^{N_t \times N_s \times d}
\end{equation}
The projected visual tokens and the embedded text tokens are then concatenated to form the multimodal input:
\begin{equation}
\mathbf{Z}_{\mathrm{all}} = [\tilde{\mathbf{Z}}_V; \mathbf{Z}_T] \in \mathbb{R}^{(N_t N_s + L_T) \times d}
\end{equation}
The concatenated sequence $\mathbf{Z}_{\mathrm{all}}$ is input into the LLM (e.g., LLaMA, Vicuna), which performs autoregressive generation for downstream tasks such as question answering:
\begin{equation}
p(\mathbf{y}|\mathbf{I_{vid}}, X_T) = \prod_{i = 1}^{L_{\text{out}}} p_\theta(y_i\mid \mathbf{Z}_{\mathrm{all}}, y_{<i})
\end{equation}
where $\mathbf{y}$ is the output response, and $\theta$ are the LLM parameters.

\subsection{Primary Objective}

In MLLMs, efficiently processing visual inputs is essential to balancing computational resources with model performance. The inherent high dimensionality and redundancy in video data pose challenges in terms of both computation and memory. Our primary objective is to formulate a token reduction strategy that minimizes the visual token count, thereby enhancing inference speed, while ensuring that the model's output quality remains nearly unchanged.

The optimization goal can be framed as minimizing the number of tokens $\mathbf{Z}_V^*$ selected from the original set $\mathbf{Z}_V$, while maintaining reasoning accuracy:
\begin{equation}
\min_{\mathbf{Z}_V^*} \quad |\mathbf{Z}_V^*| \quad \text{subject to}\, \mathbb{D}(\mathbf{y}, \mathbf{y}^*) \approx 0 \
\end{equation}
where $\mathbb{D}(\mathbf{y}, \mathbf{y}^*)$ represents a measure of divergence or difference between the original model output $\mathbf{y}$ and the output after compression $\mathbf{y}^*$.

By strategically reducing tokens along both temporal and spatial axes, our method enhances inference speed and efficiency without sacrificing the model's ability to accurately reason over visual and textual data. This makes our approach well-suited for deployment in resource-constrained scenarios where maintaining high performance at reduced computational cost is crucial.

\section{Method}

\begin{figure*}[t]
\centering
\includegraphics[scale=0.6]{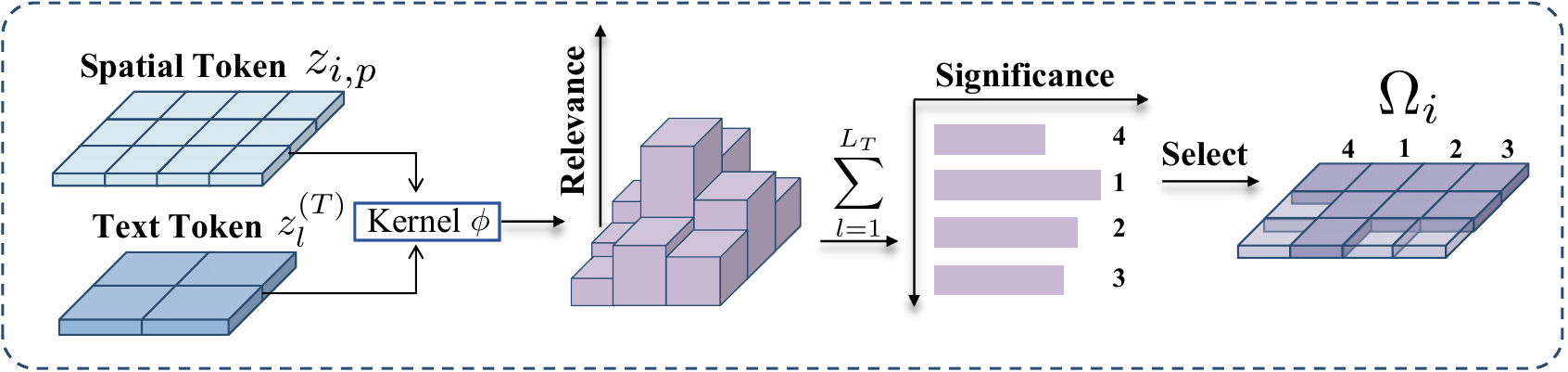}
\caption{Text-based spatial significance weights calculation. Based on the text prompt, we calculate the correlation between each visual token and the text instruction, and identify the most significant tokens.}
\label{spatial-text}
\end{figure*}

To address the challenge of excessive inference time and the degradation of model performance caused by an overabundance of prefixed visual tokens, we formalize the problem of efficient visual token compression for MLLMs as information-preserving compression and significant token selection task over the temporal and spatial axes of image/video token representations, respectively. As shown in Figure.~\ref{LLM}, our framework LFTR is entirely model-agnostic, training-free, and leverages advanced theoretical tools for principled token selection.

\subsection{Temporal Token Reduction}

Temporal redundancy is prevalent in video modality, especially at the token level for adjacent frames. We approach temporal token compression from the perspective of conditional information gain and change-point detection, aiming to adaptively reduce redundant tokens while preserving essential visual dynamics.

For each spatial location $p$, the conditional information of frame $i$ given frame $i-1$ can be formulated as:
\begin{equation}
\mathbb{I}(\mathbf{z}_{i,p};\mathbf{z}_{i-1,p}) = H(\mathbf{z}_{i,p}) - H(\mathbf{z}_{i,p}|\mathbf{z}_{i-1,p})
\end{equation}
where $H(\cdot)$ denotes differential entropy and $\mathbf{z}_{i,p}$ is the token at position $p$ in frame $i$. In practice, estimating information-theoretic quantities directly is intractable for high-dimensional features. Therefore, we employ robust proxies to quantify temporal significancy between tokens.

\begin{figure}[h]
\centering
\includegraphics[scale=0.38]{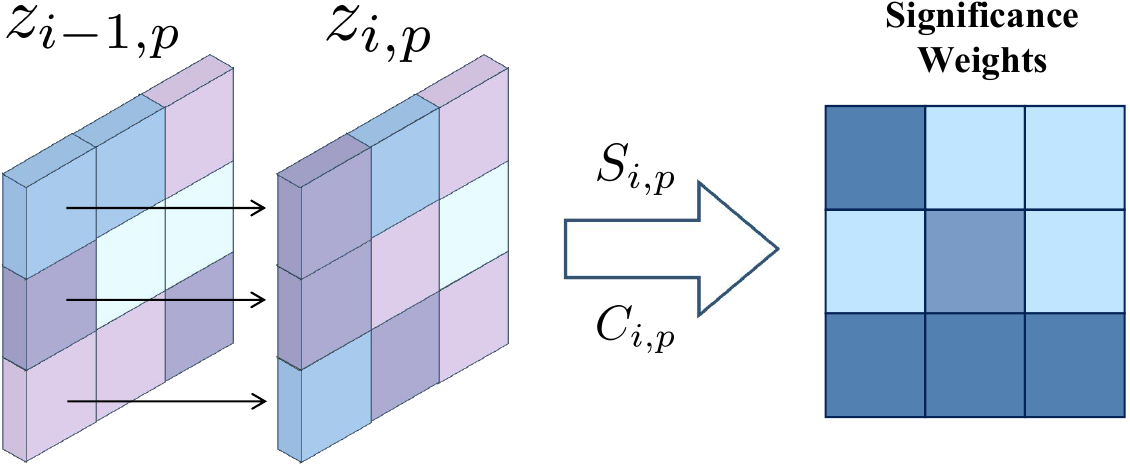}

\caption{Temporal Dimension Token Reduction Method. The significance of each spatial patch token is calculated based on its similarity or difference with the corresponding token at the same spatial position in the previous temporal frame.}
\label{temporal}
\end{figure}

We consider two practical metrics for measuring the temporal difference (or similarity) between tokens at the same spatial location across adjacent frames as shown in Figure.~\ref{temporal}.

\textbf{Summation-based Difference ($L_1$ Distance)} 
The element-wise absolute difference is computed between $\mathbf{z}_{i,p}$ and $\mathbf{z}_{i-1,p}$:
\begin{equation}
S_{i,p}  = ||\mathbf{z}_{i,p} - \mathbf{z}_{i-1,p}||_1
\end{equation}

\textbf{Cosine Similarity}  
The angular similarity is measured as:
\begin{equation}
C_{i,p} = \frac{\mathbf{z}_{i,p} \cdot \mathbf{z}_{i-1,p}}{\|\mathbf{z}_{i,p}\|_2 \|\mathbf{z}_{i-1,p}\|_2}
\end{equation}
with $\cdot$ representing the dot product. Higher values indicate greater similarity.

Instead of retaining tokens with large inter-frame differences, we focus on merging temporally redundant tokens—specifically, those that are highly similar to their temporal counterparts. The merge process is triggered based on either metric: 
For the $L_1$ difference, if $S_{i,p}$ is smaller than a predefined threshold $\tau_1$, we regard $\mathbf{z}_{i,p}$ as redundant and merge it with $\mathbf{z}_{i-1,p}$.
For cosine similarity, if $C_{i,p}$ exceeds a threshold $\tau_2$, we also perform merging.

Formally, the merge indicator $M_{i,p}$ is defined as:
\begin{equation}
M_{i,p}^{(L_1)} =
\begin{cases}
1, & \text{if } S_{i,p} < \tau_1 \\
0, & \text{otherwise}
\end{cases}
\qquad
M_{i,p}^{(\cos)} =
\begin{cases}
1, & \text{if } C_{i,p} > \tau_2 \\
0, & \text{otherwise}
\end{cases}
\end{equation}
where $M_{i,p} = 1$ indicates that $\mathbf{z}_{i,p}$ will be merged with $\mathbf{z}_{i-1,p}$ (e.g., vector averaging), and $M_{i,p}=0$ means $\mathbf{z}_{i,p}$ is retained as a new, informative token.

In this way, only tokens corresponding to actual visual changes are explicitly retained, while redundant tokens are compressed via merging. This adaptive strategy reduces both computational and memory overhead, without sacrificing the model’s ability to capture essential temporal dynamics.

\subsection{Spatial Token Compression}

Spatial redundancy among visual tokens is common, as many spatial patches offer limited value to downstream comprehension—particularly in tasks guided by text prompts. To address this, as shown in Figure.~\ref{spatial-text}, we focus primarily on a text-conditional spatial token compression strategy, while retaining topic-guided compression as a complementary approach.

\textbf{Text-Based Significance Weights.}
In MLLMs, spatial patch tokens and text prompt tokens are typically projected into a common feature space (e.g., via a multimodal projector). This enables direct assessment of the relevance between each spatial token and the input text prompt. For each frame $i$, let $\mathbf{z}_{i,p}$ ($p=1,\ldots,N_s$) denote a projected visual token, and $\mathbf{z}_l^{(T)}$ ($l=1,\ldots, L_T$) denote embedded text prompt tokens.

We calculate the text-conditioned relevance of each spatial patch token as
\begin{equation}
R_{i,p} = \sum_{l=1}^{L_T}  \phi(\mathbf{z}_{i,p}, \mathbf{z}_l^{(T)})
\end{equation}
where $\phi(\cdot,\cdot)$ denotes a correlation kernel function.

Given a spatial token reduction ratio $r_p$, we select the top $(1 - r_p) \times N_s$ tokens with the highest relevance $R_{i,p}$, pruning the remainder. Formally,
\begin{equation}
\Omega_i = \operatorname{TopK}_{p}(R_{i,p}, k = (1-r_p)N_s)
\end{equation}
\begin{equation}
\mathcal{Z}_{i}^{*} = \{\mathbf{z}_{i,p} \mid p \in \Omega_i \}
\end{equation}
This process is performed independently for each frame and can be efficiently parallelized.

\textbf{Topic-Based Significance Weights.}
For scenarios where no explicit text prompt is available, or as an auxiliary cue, we additionally propose topic-based spatial token compression. Here, we use the class token derived from the visual encoder (e.g., CLIP-ViT) as an indicator of frame-level, topic-related information. We compute the significance weight of each spatial token via its cosine similarity with the [CLS] token:

\begin{equation}
R_{i,p} = \langle \hat{\mathbf{z}}_{i,\text{[CLS]}}, \hat{\mathbf{z}}_{i,p} \rangle
\end{equation}
where $\hat{\cdot}$ denotes normalization.
Analogous to the text-based approach, we retain the top $(1 - r_p) \times N_s$ tokens with the highest topic relevance per frame. This strategy is lightweight and complementary, providing an unsupervised method to filter uninformative patches.

\subsection{Joint Compression}

We integrate the temporal and spatial token compression modules in a unified framework, with each stage tailored to minimize redundancy and preserve task-relevant information.
Given a video represented as token tensor $\mathbf{Z}_V = \{\mathbf{z}_{i,p}\}$, we first reduce temporal redundancy by merging spatial tokens across frames whenever temporal similarity indicates negligible change:
\begin{equation}
\mathbf{z}_{i,p} \leftarrow \begin{cases}
\operatorname{Merge}(\mathbf{z}_{i-1,p},\, \mathbf{z}_{i,p}), & \text{if}\; S_{i,p} < \tau_1\ (\text{or}\ C_{i,p} > \tau_2) \\
\mathbf{z}_{i,p}, & \text{otherwise}
\end{cases}
\end{equation}
where $S_{i,p}$ and $C_{i,p}$ denote the summation-based difference and cosine similarity described previously, respectively. This process adaptively merges temporally redundant tokens, effectively reducing the number of distinct temporal tokens.

Next, for each retained frame, we apply spatial token selection. In text-conditioned scenarios, we compute the text-conditioned relevance score for each spatial token (as detailed previously) and keep the top-$K_s$ tokens:
\begin{equation}
\mathcal{S}_i^* = \operatorname{TopK}_p\left(R_{i,p},\, K_s\right)
\end{equation}
where $R_{i,p}$ is the spatial significance weight.

The final compressed visual token set is:
\begin{equation}
\mathbf{Z}_V^* = \{\mathbf{z}_{i,p}\;\vert\; i \in \mathcal{T}^*,\, p \in \mathcal{S}_i^*\}
\end{equation}
where $\mathcal{T}^*$ is the set of effective temporal indices after merge-based compression, and $\mathcal{S}_i^*$ is the set of selected spatial patch indices for frame $i$.

This joint pipeline can be understood as an adaptive filtering process, aiming to preserve the most salient and task-relevant tokens while adhering to computational constraints. Formally, the token selection can be interpreted as maximizing the conditional mutual information between the compressed visual token set and the text prompt:
\begin{equation}
\max_{\mathcal{T}^*,\{\mathcal{S}_i^*\}}\, \mathbb{I}\left(\mathbf{Z}_V^*;\, \mathbf{Z}_T\right)
\end{equation}
subject to token budget or resource limits.

In summary, our strategy exploits temporal redundancy by merging highly similar tokens across consecutive frames, and then selectively retains spatial tokens most relevant to the text prompt (or, optionally, the visual topic). This two-stage process enables scalable and adaptive token reduction, facilitating efficient visual-language modeling. A key strength of our approach lies in its fully structure-agnostic design, which ensures broad compatibility with diverse MLLM architectures. Unlike prior methods such as LLaVA-PruMerge and freePruner, our method does not rely on specific architectural components such as vision encoders or class tokens, making it readily applicable across a wide range of models.

\begin{table*}[h]
\centering
\small
\setlength{\tabcolsep}{5pt}
\renewcommand{\arraystretch}{1.2}
\caption{Performance comparison of our method integrated into various MLLMs on video reasoning benchmarks, reported in terms of accuracy (\%) and core. We follow the evaluation protocol of Video-LLaVA and utilize ChatGPT-Assistant for performance assessment.}
\label{tab:video-qa}
\resizebox{\textwidth}{!}{
\begin{tabular}{lcccccccccc}
\toprule
\textbf{Methods} & \textbf{LLM size} & \multicolumn{2}{c}{\textbf{MSVD-QA}} & \multicolumn{2}{c}{\textbf{MSRVTT-QA}} & \multicolumn{2}{c}{\textbf{ActivityNet-QA}} & \multicolumn{2}{c}{\textbf{TGIF-QA}}\\
 &  & \textbf{Accuracy} & \textbf{Score} & \textbf{Accuracy} & \textbf{Score} & \textbf{Accuracy} & \textbf{Score} & \textbf{Accuracy} & \textbf{Score}\\
\midrule
FrozenBiLM & 1B & 32.2 & - & 16.8 & - & 24.7 & - &41.0 & -\\
VideoChat & 7B & 56.3 & 2.8 & 45.0 & 2.5 & -- & 2.2 &34.4 &2.3\\
LLaMA-Adapter & 7B & 54.9 & 3.1 & 43.8 & 2.7 & 34.2 & 2.7 & - & -\\
Video-ChatGPT & 7B & 64.9 & 3.3 & 49.3 & 2.8 & 35.2 & 2.7 &51.4 & 3.0\\
\midrule
MiniGPT4-Video & 7B & 72.9 & 3.8 & 58.8 & 3.3  & 45.8 &3.2 & 67.9 & 3.7\\
\rowcolor{lightgray}MiniGPT4-Video + LFTR ($8\times$) & 7B & 75.2 & 4.0 & 61.0 & 3.4 & 46.0 & 3.4 & 70.5 &3.7 \\

\midrule
Video-LLaVA & 7B & 70.7 & 3.9 & 59.2 & 3.5 & 45.3 & 3.3  &70.0 & 4.0\\
Video-LLaVA + PruMerge ($8\times$) & 7B & 71.1 & 3.9 & 58.4 & 3.5 & 48.3 & 3.4 & - & - \\
Video-LLaVA + freePruner ($8\times$) & 7B & 71.3 & 3.9 & 59.5 & 3.5 & 48.4 &3.4 & - & -\\
\rowcolor{lightgray} Video-LLava + LFTR ($8\times$)  & 7B & 77.2 & 4.2 & 64.7 & 3.8 & 50.4 & 3.5 &76.4 &4.5 \\
\rowcolor{lightgray} Video-LLava + LFTR ($16\times$) & 7B & 72.7 & 4.0 & 61.4 & 3.7 & 48.5 & 3.6 & 70.5 & 4.2 \\
\midrule
QwenVL-2.5 & 3B & 80.3 & 4.1 & 62.6 & 3.6  & 81.5 &4.2 & 78.5 & 4.1\\
\rowcolor{lightgray}QwenVL-2.5+ LFTR ($8\times$) & 3B & 81.9 & 4.1 & 63.2 & 3.8  & 81.6 &4.4 & 78.9 & 4.2\\
QwenVL-2.5 & 7B & 83.3 & 4.1 & 63.9 & 3.7  & 82.4 &4.3 & 79.9 & 4.1\\
\rowcolor{lightgray}QwenVL-2.5+ LFTR ($8\times$) & 7B & 84.6 & 4.2 & 66.4 & 3.9  & 83.9 &4.4 & 81.1 & 4.3\\
\bottomrule
\end{tabular}
}
\end{table*}
\section{Experiments}

In this section, we first integrate our visual token compression method into the open source Qwen2.5-VL~\cite{bai2025qwen2}, Video-LLaVA~\cite{lin2023video} and MiniGPT4-Video~\cite{ataallah2024minigpt4} frameworks. We then evaluate the efficiency improvement brought by our LFTR approach, including its impact on LLM inference and runtime overhead as a plug-and-play module. To demonstrate the flexibility and robustness of our token reduction strategies, we examine the performance of different combinations of spatial and temporal reduction strategies as well as ablation studies to assess the effectiveness of the proposed spatial and temporal compression techniques under varying token reduction ratios.

\subsection{Zero-shot Evaluation on Video-LLM}
\label{expsec1}

We apply our token reduction method to Video-LLaVA, MiniGPT4-Video, and QwenVL-2.5. The thresholds $\tau_1$ and $\tau_2$ are set adaptively based on the input values to maintain approximately constant compression rates. Under this configuration, our method achieves visual token retention rates as low as 12.5\% and 6.25\% of the original input—corresponding to $8\times$ and $16\times$ reduction ratios, respectively—without requiring any additional training or fine-tuning.
 We evaluate the zero-shot inference ability of MLLMs with LFTR on four video question-answering benchmarks: MSVD-QA~\cite{chen2023x}, MSRVTT-QA~\cite{xu2016msr}, TGIF-QA~\cite{jang2017tgif} and ActivityNet-QA~\cite{yu2019activitynet}. We report the accuracy and score, which is assessed using GPT-3.5-Turbo.

The results of this integration are presented in Table~\ref{tab:video-qa}. We compare our approach with two SOTA token reduction methods: LLaVA-PruMerge~\cite{shang2024llava} and freePruner~\cite{xu2024freepruner}. Our method outperforms both baselines, even under more aggressive token reduction settings. These results demonstrate that our approach not only substantially reduces the number of visual tokens—thereby improving inference efficiency—but also enhances the overall performance of Video-LLMs. Moreover, its successful application across three distinct model architectures and processing workflows underscores the robustness and plug-and-play compatibility of our framework.

Importantly, our method can be readily integrated into widely adopted MLLM architectures without imposing constraints on the vision encoder design (as required by methods like freePruner) or relying on specific components such as the class token (as in PruMerge). This architecture-agnostic property significantly broadens the applicability of our approach.

\subsection{Efficiency Analysis}
\label{expsec2}

To elucidate the computational efficiency gains enabled by LFTR, we adopt the roofline-based LLM Viewer framework~\cite{yuan2024llm} to analyze its theoretical impact on LLM processing efficiency.  Our investigation is grounded in a theoretical scenario tailored to highlight the impact of LFTR on processing efficiency within LMMs. Consider a typical scenario where an image of dimensions $224\times 224$ pixels is processed using a CLIP-ViT model with 8 video frames sampled from the video inputs, resulting in a total amount of $2056$. Accompanying this image is a text prompt, assumed to contain 50 tokens for the sake of this analysis. Following the evaluation settings in LLaVA-PruMerge~\cite{shang2024llava}, we assess the performance under FP16 and INT4 quantization on an NVIDIA A6000 GPU. The results demonstrate that by reducing visual tokens to only 6.25\% of the original input without any training. LFTR yields substantial improvements in prefill time, accessing memory, and activation storage. Given that the deployment of Video-LLMs is often constrained by memory consumption for activation/KV-cache and response latency, our method shows strong potential for practical, real-world applications.
In addition, we report the average inference time over 13,000 Video-QA samples under the same experimental settings. We evaluate various token reduction ratios and their corresponding impact on inference time. As shown in Table~\ref{tab:avg_inf_t}, the token reduction significantly enhances MLLM inference efficiency while simultaneously improving model performance. The results demonstrate that token reduction process can contribute to a significant efficiency improvement in real application settings. 
\begin{table*}[h]
\centering
\small
\setlength{\tabcolsep}{3pt}
\renewcommand{\arraystretch}{1.2}
\caption{Computation Cost Analysis.}
\label{tab:efficiency}
\begin{tabular}{lccccccc}
\toprule
\textbf{Method} & \textbf{LLM Backbone} & \textbf{Quantization} & \textbf{OPs (TB)} & \textbf{Prefill Time (ms)} & \textbf{Accessing Memory (GB)} & \textbf{Storing Activation (GB)} \\
\midrule
LLaVA & Vicuna-7B & FP16 & 29.6 & 233.3 & 67.3 & 25.6 \\
\rowcolor{lightgray} LLaVA w/ \texttt{LFTR} & Vicuna-7B & FP16 & 2.3 & 19.5 & 15.0 & 0.8 \\
LLaVA & Vicuna-7B & INT4 & 29.6 & 102.3 & 16.8 & 6.4 \\
\rowcolor{lightgray} LLaVA w/ \texttt{LFTR} & Vicuna-7B & INT4 & 2.3 & 7.8 & 3.7 & 0.2 \\

\bottomrule
\end{tabular}
\end{table*}
\begin{table}[h]
\setlength{\tabcolsep}{8pt}
\centering
\caption{MLLM Inference Time Consumption and Accuracy (\%)}
\label{tab:avg_inf_t}
\begin{tabular}{lccccc}
\toprule
\textbf{Ratio} & \textbf{$1\times$}& \textbf{$2\times$} & \textbf{$4\times$} & \textbf{$8\times$} & \textbf{$16\times$}\\
\midrule
Avg Time(s) & 2.0 & 1.6 & 1.3 & 1.2 & 1.2\\
Accuracy & 70.7 & 75.4 & 75.9 & 77.2 &  72.7 \\
\bottomrule
\end{tabular}
\end{table}

\begin{table}[t]
\centering
\caption{Comprehensive ablation and combination study: accuracy (\%). Each block shows results for a spatial reduction method at various rates, under different temporal reduction strategies and rates. }
\label{tab:comprehensive_ablation}
\begin{tabular}{@{}c|c|ccc|ccc@{}}
\toprule
\multirow{2}{*}{\textbf{}} & \multirow{2}{*}{\textbf{Spatial}} & \multicolumn{3}{c|}{\textbf{CS Temporal}} & \multicolumn{3}{c}{\textbf{SD Temporal}} \\
& & $1\times$ & $2\times$ & $4\times$ & $1\times$ & $2\times$ & $4\times$ \\
\midrule
\multirow{3}{*}{\textbf{Text}} 
 & $1\times$ & 70.6 & 75.0 & 76.3 & 70.6 & 75.1 & 76.1 \\
 & $2\times$ & 71.3 & 75.9 & 77.1 & 71.3 & 72.2 & 77.2 \\
 & $4\times$ & 71.9 & 70.7 & 71.4 & 71.9 & 76.4 & 73.6 \\
\midrule
\multirow{3}{*}{\textbf{Topic}} 
 & $1\times$ & 70.6 & 75.0 & 76.3 & 70.6 & 75.1 & 76.1 \\
 & $2\times$ & 71.0 & 74.8 & 74.6 & 71.0 & 66.2 & 70.1 \\
 & $4\times$ & 68.7 & 71.1 & 65.0 & 68.7 & 65.6 & 63.3 \\
\bottomrule
\end{tabular}
\end{table}

\subsection{Ablation Study}
\label{expsec:token-eval}

We conduct a thorough ablation study to evaluate our proposed token reduction strategies with the Video-LLaVA architecture. Both joint and individual applications are evaluated, with a focus on inference accuracy and efficiency.

\textbf{Experimental Setup.}
All experiments employ the MSVD-VQA benchmark, with inference accuracy measured via the ChatGPT. We evaluate summation-based difference (\textbf{SD}) and cosine similarity (\textbf{CS}) temporal reduction methods as well as text-based and topic-based spatial reduction methods, each at $2\times$ and $4\times$ rates. Both independent (single-dimension with  $1\times$ of another dimension) and joint (combined) reductions are assessed.

\textbf{Results and Analysis.}
Table~\ref{tab:comprehensive_ablation} presents a unified view of performance across all configurations. The rows denote spatial token reduction methods and rates, while the columns correspond to temporal reduction methods and rates. The entries along the top row and leftmost column, corresponding to cases where either the spatial or temporal reduction rate is set to $1\times$, serve as ablation studies of single-dimensional reduction, facilitating direct comparison with joint reduction strategies.

\textbf{Single-Dimension Reduction.}
Analyzing Table~\ref{tab:comprehensive_ablation}, where either the spatial or temporal reduction rate is fixed at $1\times$, we observe that both spatial and temporal token reductions, when applied independently, preserve strong inference accuracy even at reduction rates as high as $4\times$. \textbf{SD} temporal reduction and text-based spatial reduction typically outperform their counterparts, confirming the importance of both temporal event selection and prompt-aligned spatial focus.

\textbf{Joint Reduction Effects and Strategy Comparison.}
Moving to joint application of temporal and spatial token strategies, joint reduction strategies show that Combining both dimensions allows for more aggressive token reduction (up to $16\times$) with only moderate accuracy degradation. The pairing of summation-based temporal and text-based spatial reduction offers the best trade-off, with $4\times$ reduction in both yielding 73.6\%—comparable to or better than some single-dimension reductions at lower rates.

In summary, our ablation and joint analysis demonstrate the effectiveness of our token reduction strategies. Both individual and combined reductions deliver significant efficiency gains while maintaining competitive or even superior accuracy, enabling scalable deployment of large-scale multimodal models.

\section{Conclusion}

In this work, we propose a learning-free token reduction (LFTR) method for multi-modal large language models (MLLMs). Our approach can be seamlessly integrated into a wide range of MLLM architectures in a plug-and-play fashion, requiring no additional training or fine-tuning. It effectively reduces token counts across both temporal and spatial dimensions for video inputs, thereby enhancing inference speed and computational efficiency without compromising the model’s capacity for accurate visual-textual reasoning.

We conduct extensive experiments to validate the effectiveness of LFTR. Specifically, we evaluate three distinct MLLM architectures on zero-shot Video-QA benchmarks. The results demonstrate that LFTR consistently improves performance, even under aggressive token reduction rates of up to $16\times$. For efficiency analysis, we quantify the computational savings at the LLM stage, confirming the significant inference acceleration achieved by our method. Additionally, we examine various combinations of token reduction strategies and perform comprehensive ablation studies to assess the robustness and consistency of our approach. Notably, LFTR is lightweight and easily deployable, allowing seamless integration into popular MLLM architectures. Its orthogonality to existing acceleration techniques further ensures broad compatibility across deployment scenarios.

\bibliography{ref_aaai}

\end{document}